\pdfoutput=1
\documentclass[letterpaper,10pt,conference]{IEEEtran}  

\IEEEoverridecommandlockouts                              

\usepackage{graphicx}
\usepackage{tabularx}
\usepackage{url}
\usepackage{caption}
\usepackage[table]{xcolor}
\usepackage{xcolor}
\usepackage{soul}
\usepackage{courier}
\usepackage[T1]{fontenc}
\usepackage{epstopdf}
\newcolumntype{Y}{>{\centering\arraybackslash}X}

\title{\LARGE \bf
Using Socially Expressive Mixed Reality Arms\protect\\ for Enhancing Low-Expressivity Robots
}

\author{Thomas Groechel, Zhonghao Shi, Roxanna Pakkar, and Maja J Matari\'c 
\thanks{This work was supported by the NSF Expeditions in Computing Grant for Socially Assistive Robotics (NSF IIS-1129148).}
\thanks{Thomas Groechel, Zhonghao Shi, Roxanna Pakkar, and Maja J Matari\'c are all
with the Interaction Lab, Department of Computer Science,
University of Southern California, Los Angeles, CA 90089, USA\protect\\
\texttt{\{groechel,zhonghas,pakkar,mataric\}@usc.edu}}%
}

\begin{document}



\maketitle

\thispagestyle{empty}
\pagestyle{empty}

\begin{abstract}

Expressivity--the use of multiple modalities to convey internal state and intent of a robot--is critical for interaction. Yet, due to cost, safety, and other constraints, many robots lack high degrees of physical expressivity. This paper explores using mixed reality to enhance a robot with limited expressivity by adding virtual arms that extend the robot's expressiveness. The arms, capable of a range of non-physically-constrained gestures, were evaluated in a between-subject study ($n=34$) where participants engaged in a mixed reality mathematics task with a socially assistive robot. The study results indicate that the virtual arms added a higher degree of perceived emotion, helpfulness, and physical presence to the robot. Users who reported a higher perceived physical presence also found the robot to have a higher degree of social presence, ease of use, usefulness, and had a positive attitude toward using the robot with mixed reality. The results also demonstrate the users' ability to distinguish the virtual gestures' valence and intent.
\end{abstract}

\section{Introduction}

Socially assistive robots (SAR) have been shown to have positive impacts in a variety of domains, from stroke rehabilitation \cite{mataric2007socially} to tutoring \cite{clabaugh2017predicting}. Such robots, however, typically have low-expressivity due to many physical, cost, and safety constraints. \textit{Expressivity} in  Human-Robot Interaction (HRI) refers to the robot's ability to use its modalities to non-verbally communicate the robot's intentions or its internal state \cite{charisi2019expressivity}. Higher levels of expressiveness have been shown to increase trust, disclosure, and companionship with a robot \cite{martelaro2016tell}. Expressivity can be conveyed with dynamic actuators (e.g., motors) as well as static ones (e.g., screens, LEDs) \cite{balit2018pear}.  HRI research into gesture has explored head and arm gestures, but many nonhumanoid robots partially or completely lack those features, resulting in low social expressivity \cite{cha2018survey}.

\textit{Social expressivity} refers to expressions related to communication of affect or emotion. In social and socially assistive robotics, social expressivity has been used for interactions such as expressing the robots emotional state through human-like facial expressions \cite{chen2018reverse, meghdari2016spontaneous, kkedzierski2013emys}, gestures \cite{cha2018survey}, and physical robot poses \cite{bretan2015emotionally}.  In contrast, \textit{functional expressivity} refers to the robot's ability to communicate its functional capabilities (e.g., using turn signals to show driving direction). Research into robot expressiveness has explored insights from animation, design, and cognitive psychology \cite{charisi2019expressivity}.

\begin{figure}[t]
    \centering
    \includegraphics[width=\columnwidth]{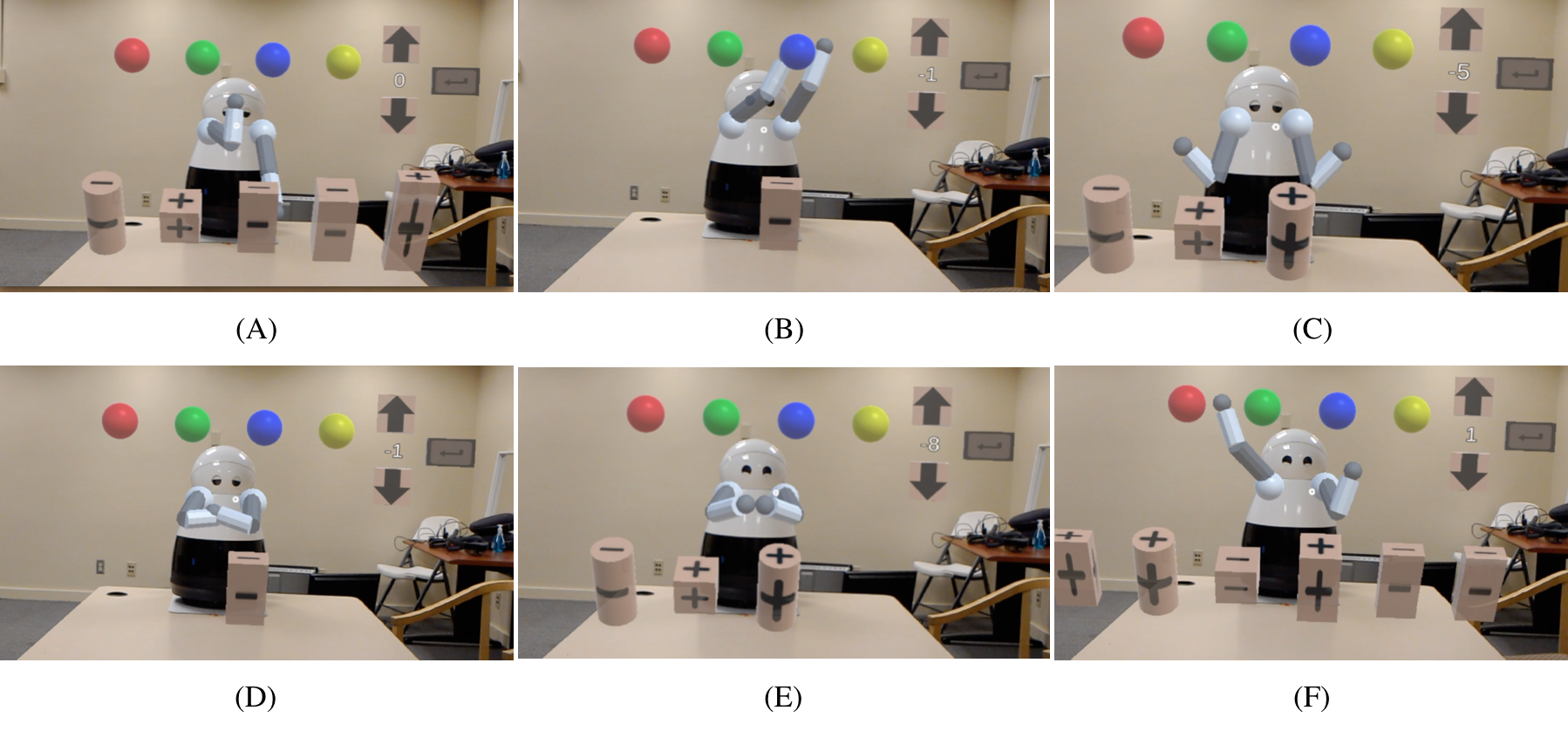}
    \captionof{figure}{This work explores how mixed reality robot extensions can enhance low-expressivity robots by adding social gestures. Six mixed reality gestures were developed: (A) facepalm, (B) cheer, (C) shoulder shrug, (D) arm cross, (E) clap, and (F) wave dance.}
    \label{fig:sixanims}
 \end{figure}
 
The importance of expressivity and the mechanical, cost, and safety constraints of physical robots call for exploring new modalities of expression, such as through the use of \textit{augmented reality (AR)} and \textit{mixed reality (MR)}. \textit{AR} refers to the ability to project virtual objects onto the real world without adherence to physical reality, while \textit{MR} refers to virtual objects projected onto the real world while respecting physical reality and reacting to it.

Using virtual modalities for robots has led to the emerging field of Augmented Reality Human-Robot Interaction (AR-HRI), which encompasses AR, MR, and virtual reality (VR) robotics. AR-HRI has already had advances in the functional expression of a robot \cite{walker2018communicating, williams2019virtual} but has not yet explored social expressiveness. Introducing such expressiveness into AR-HRI allows us to leverage the positive aspects of physical robots--embodiment and physical affordances \cite{deng2019embodiment}--as well as the positive aspects of mixed reality--overcoming cost, safety, and physical constraints. {\it This works aims to synergize the combined benefits of the two fields by creating mixed reality, socially expressive arms for low social expressivity robots (Fig. \ref{fig:sixanims}).}

 This paper describes the design, implementation, and validation of MR arms for a low-expressivity physical robot. We performed a user study where participants completed a mixed reality mathematics task with a robot. This new and exploratory work in AR-HRI did not test specific hypotheses; empirical data were collected and analyzed to inform future work. The results demonstrate a higher degree of perceived robot emotion, helpfulness, and physical presence by users who experienced the mixed reality arms on the robot compared to those who did not. Participants who reported a higher physical presence also reported higher measures of robot social presence, perceived ease of use, usefulness, and had a more positive attitude toward using the robot with mixed reality. The results from the study also demonstrate consistent ratings of gesture valence and identification of gesture intent.

\section{Background and Related Work}

\subsection{Augmented Reality Human-Robot Interaction}

AR and MR can be delivered through projectors \cite{gillen2012beyond}, tablets \cite{hashimoto2011touchme}, and head-mounted displays \cite{walker2018communicating}. Projectors allow for hands-free communication, but are limited with respect to user input. In contrast, tablets allow for direct, intuitive, and consistent user input, but restrict users to a 2D screen, eliminating hands-free, kinesthetic interactions. An augmented reality head-mounted display (ARHMD) aims to remove those limitations by allowing consistent, direct input and hands-free interaction. ARHMDs, such as the Microsoft Hololens, allow for high quality, high fidelity, hands-free interaction.

The reality-virtuality continuum spans the range of technologies from physical reality, to differing forms of mixed reality, to full virtual reality \cite{milgram1995telerobotic}. Exploring that continuum for enhancing robot communication is a nascent area of research; Augmented Reality Human-Robot Interaction (AR-HRI) has been gaining attention \cite{williams2019virtual}.

Creating mixed reality experiences with robots is now possible with open-source tools \cite{RosSharp2018}. Work to date has largely focused on signalling functional intent \cite{walker2018communicating, williams2019virtual} and teleoperation \cite{zhang2018deep, hedayati2018improving, lipton2018baxter}. For example, functional signalling used in Walker et al. \cite{walker2018communicating} allowed nonexpert users to understand where a robot was going in the real world, which was especially useful for robots with limited expressive modalities. 

The use of AR for HRI, however, is a very new area of research \cite{williams2018augmented}. Specifically, research into AR-HRI's to date has focused on functional expression, with little work on social expression. Survey analysis across the Miligram virtuality continuum has shown early work in social mixed reality for robots to be limited \cite{holz2009robots}. Examples include adding a virtual expressive face to a Roomba vacum cleaning robot \cite{young2007robot} and adding a virtual avatar on a TurtleBot mobile robot \cite{dragone2006mixing}.  To the best of our knowledge, the virtual overlays have not been pursued further since the survey was conducted in 2009, leaving opportunities open for exploring socially expressive AR-HRI design.

\subsection{Social Expressivity and Presence in HRI}

Research in HRI has explored robot expressiveness extensively, including simulating human facial expressions on robots \cite{chen2018reverse, meghdari2016spontaneous, kkedzierski2013emys}, gestures \cite{cha2018survey}, and physical social robot poses \cite{bretan2015emotionally}.
Increased social expressivity has been shown to build rapport and trust \cite{martelaro2016tell}.

For socially interactive robots, social presence depends on the ability to communicate expected social behaviors through the robot's available modalities \cite{cha2018survey, oh2018systematic}. A high degree of social presence can be beneficial to user sympathy and intimacy \cite{jimenez2015learning}. This effect has been validated in various domains, including museum robots \cite{nourbakhsh1999affective} and robot math tutors \cite{clabaugh2017predicting}. Since physical robots are limited by cost, physical safety, and mechanical constraints, socially interactive robots often explore additional communication channels, ranging from lights \cite{baraka2016enhancing} to dialogue systems \cite{belpaeme2013multimodal}.  The work presented here also explores an additional communication channel, by taking advantage of the high fidelity of MR and the lack of physical constraints to evaluate the effectiveness of mixed reality gestures on increasing social expressiveness.

\section{Robot Gesture Design and Implementation}
\begin{figure}[t]
  \centering
  \includegraphics[width=\columnwidth]{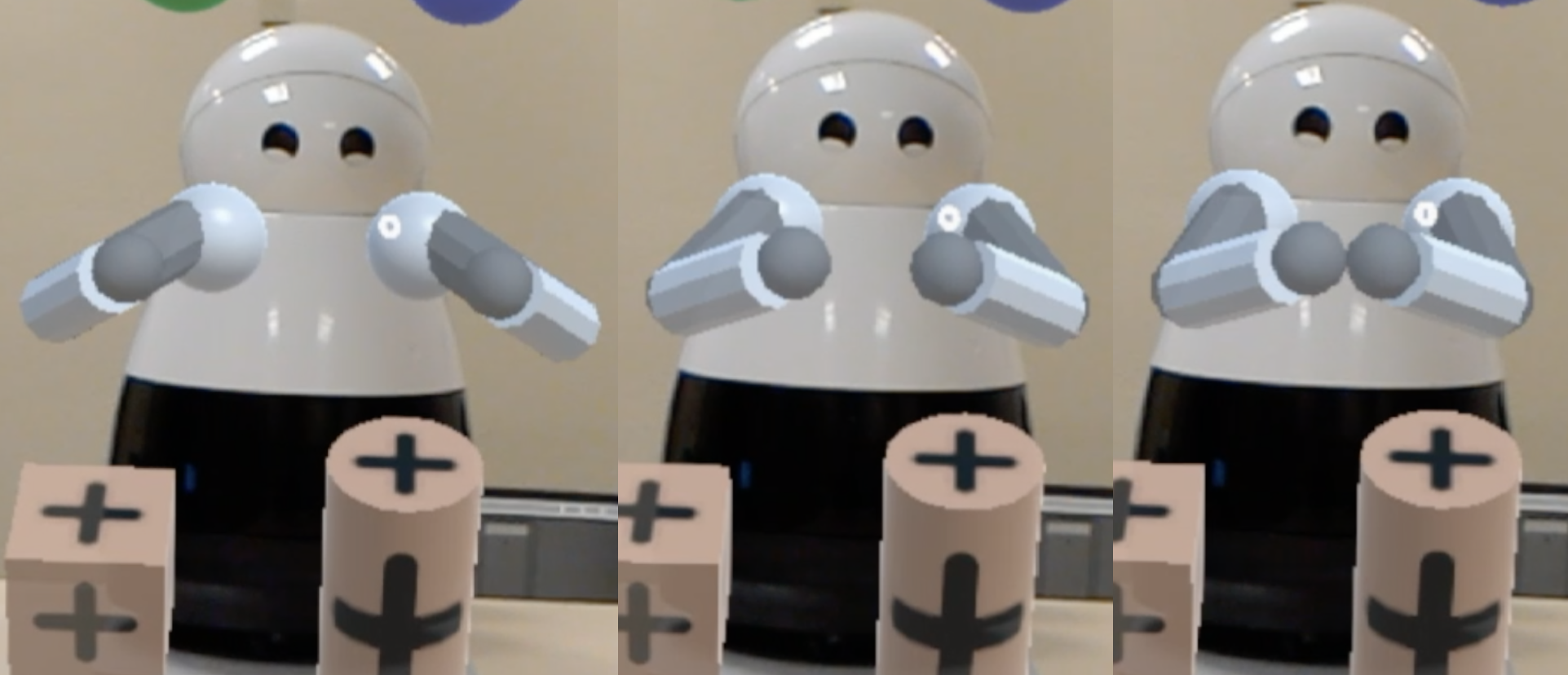}
  \caption{Keyframes for Kuri's clapping animation.}
  \label{fig:clapanim}
\end{figure}

To study MR arms, we chose a mobile robot with very low expressivieness: the Mayfield Robotics Kuri (Fig. \ref{fig:clapanim}), formerly a commercial product. Kuri is 50 cm tall, has 8 DoF (3 in the base, 3 between the base and head, and 1 in each of the two eyelids), and is equipped with an array of 4 microphones, dual speakers, lidar, chest light, and a camera behind the left eye. While very well engineered, Kuri is an ideal platform for the exploration of AR-HRI in general, and MR gestures in particular, because of its lack of arms.

\subsection{Implementation}
We used the Microsoft Hololens ARHMD, which is equipped with a $30^\circ$ x $17.5^\circ$ field of view (FOV), an IMU, a long-range depth sensor, an HD video camera, four microphones, and an ambient light sensor. We developed the mixed reality extension prototypes in Unity3D, a popular game engine. For communication between Kuri and the Hololens, we used the open source ROS\# library \cite{RosSharp2018}. The full arm and experiment implementation is open-source and available at \url{https://github.com/interaction-lab/KuriAugmentedRealityArmsPublic}.

We developed a balanced set of positive gestures (a dancing cheer, a clap, and a wave-like dance) and negative gestures (a facepalm, a shoulder shrug, and crossing the arms), as shown in Fig. \ref{fig:sixanims}. We used the Unity built-in interpolated keyframe animator to develop each gesture animation and a simple inverse kinematic solver to speed up the development of each keyframe.

\subsection{Gesture Design} \label{sssec:gesturedesign}
In designing the virtual gestures for the robot, we took inspiration from social constructs of collaboration, such as pointing to indicate desire \cite{tomasello2010origins}, and emblems and metaphoric gestures \cite{mcneill1992hand}, such clapping to show approval. The inclusion of such gestures goes beyond the current use of mostly audio and dance feedback in many socially assistive robot systems \cite{leyzberg2012physical, clabaugh2017predicting, scassellati2018improving}.

Work in HRI has explored Disney animation principles \cite{thomas1995illusion}, typically either in simulation or with physically constrained robots \cite{Takayama2011,Gielniak2012,Ribeiro2012}.  In this work, we explored a subset of Disney principles--squash and stretch, exaggeration, and staging--in the context of MR arm gestures. Each principle was considered for its benefit over physical world constraints. Squash and stretch gives flexibility and life to animations bringing life to robots that are rigid. Exaggeration has been shown to aid robot social communication \cite{Gielniak2012}. Staging was considered for its role in referencing objects using arms to allow for joint attention.

Informed by feedback from a pilot study we conducted, the animated gestures were accompanied by physical body expressions to make the arms appear integrated with Kuri. For positive gestures, Kuri performed a built-in happy gesture (named ``gotit'') that involved the robot's head moving up and emitting a happy, rising tone. For negative gestures, Kuri performed a built-in sad gesture (named ``sad'') that involved the robot's head moving down and being silent.

We also explored the use of deictic (i.e., pointing) gestures; these have been recently explored in AR-HRI but only through a video survey \cite{williams2019virtual}.  The gestures had both functional and social purposes, as discussed in Section \ref{sssec::task2}.

\section{Experiment Design}
A single-session experiment consisting of two parts was conducted with approval by our university IRB (UP-16-00603). The first part was a two-condition between-subjects experiment to test the mixed reality arms. All participants wore the ARHMD and interacted with both physical and virtual objects as part of a math puzzle game. The independent variable was whether participants had arms on their Kuri robot (Experiment condition) or not (Control condition). We collected subjective measures including perceived physical presence, social presence, ease of use, helpfulness, and usefulness from Heerink et al. \cite{heerink2010assessing}, adapted for the mixed reality robot. Task efficiency was objectively measured using completion time as is standard in AR-HRI \cite{walker2018communicating}. After the first part of the experiment was completed, a survey of 7-point Likert scale questions abd a semi-structured interview were administered. 

The second part of the experiment involved all participants in a single condition. The participants were shown a video of each of the six MR arm gestures and asked to rate each gesture's valence on a decimal scale from very negative (-1.00) to very positive (+1.00), as in prior work \cite{marmpena9does}, and to describe verbally, in written form, what each gesture conveyed. 

\subsection{Part 1: Mixed Reality Mathematics Puzzles}
\begin{figure}[t]
  \centering
  \includegraphics[width=\columnwidth]{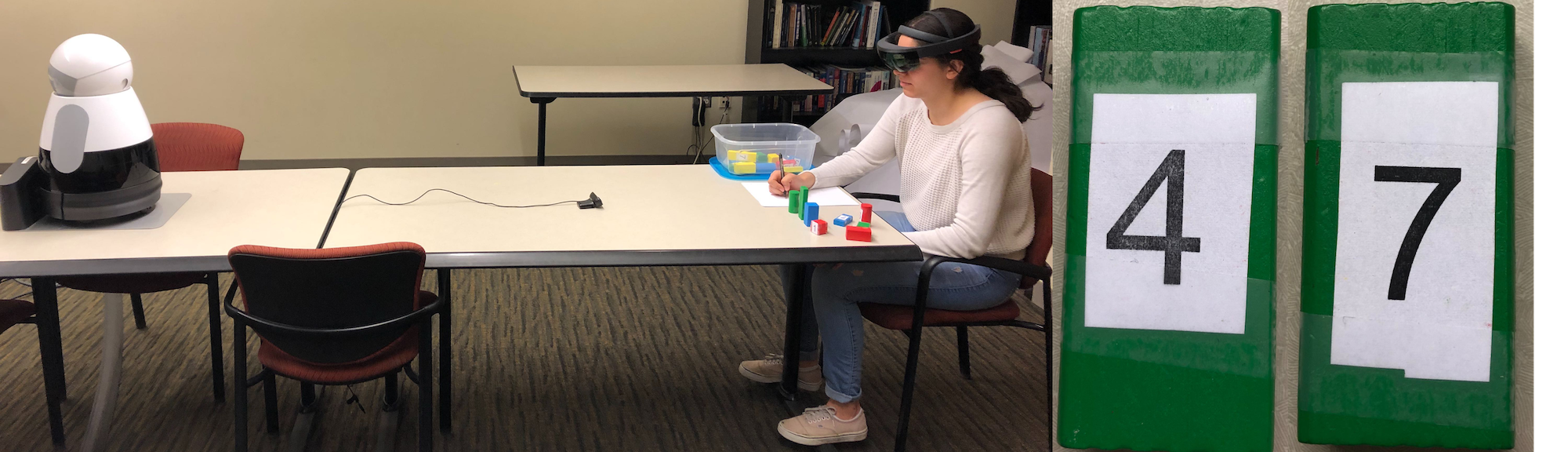}
  \caption{Participant wearing the Hololens across from Kuri (left). Two sides of a single physical cuboid block (right).}
  \label{fig:twosideblocks}
\end{figure}

\begin{figure}[b]
  \centering
  \includegraphics[width=\columnwidth]{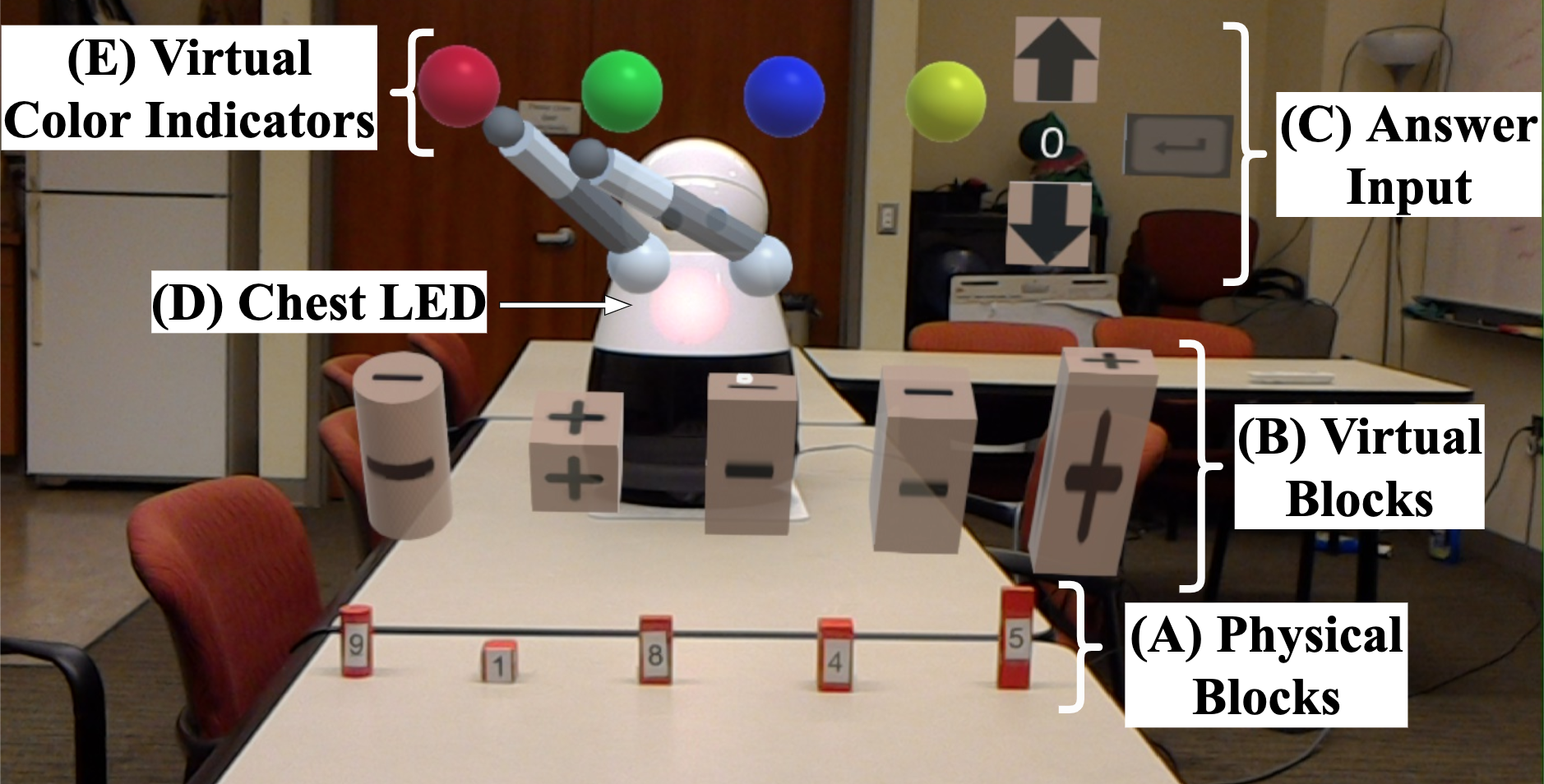}
  \caption{View when clicking a virtual block. Kuri is displaying red on its chest and pointing to the red sphere to indicate the virtual clicked block to the corresponding physical block color. From left to right, the blocks read: 9, 1, 8, 4, 5.}
  \label{fig:environment}
\end{figure}

Participants wore the Hololens and were seated across from Kuri (Fig. \ref{fig:twosideblocks}) 
with a set of 20 colored physical blocks on the table in front of them. The blocks were numbered 1-9. The block shapes were: cylinder, cube, cuboid, wide cuboid, and long cuboid.  The block colors were: red, green, blue, and yellow. The participants' view from the Hololens can be seen in Fig. \ref{fig:environment}, with labels for all objects pertinent to solving the mathematics puzzle. The view included cream-colored blocks in the same variety of shapes, labeled with either a plus (+) or minus (-) sign.  Participants were asked to solve an addition/subtraction equation based on information provided on the physical and virtual blocks, and virtually input the numeric answer. 

Participants were shown anywhere from 1 to 8 cream-colored \textbf{virtual blocks (B)} for each puzzle. To discover the hidden virtual block color, participants clicked on a virtual block (by moving the Hololens cursor over it and pressing a hand-held clicker); in response, Kuri's \textbf{chest LED (D)} lit up in the hidden block color. In the Experiment condition, Kuri also used MR arms to point to the \textbf{virtual color indicator (E)} of that color.

Once the color was so indicated, the participants selected a \textbf{physical block (A)} with the same shape and color. The number displayed on the physical block was part of the math equation. The + or - on the virtual block indicated whether the number should be added or subtracted. Once all virtual-to-physical correspondences for the blocks were found, participants added or subtracted the numbers into a running sum (initialized at 0), calculated the final answer, and input it into the virtual \textbf{answer input (C)}.

At the start of the session, participants were guided through the process in a scripted tutorial (Fig. \ref{fig:environment}). They were told to click on the virtual cylinder on the far left. Once clicked, Kuri lit up its chest LED in red and pointed at the red virtual ball-shaped color indicator. Participants then grabbed the red cylinder with the number 9 on it. This process was repeated for all the blocks. The resulting sum was: \{(-9), (+1), (-8), (-4), (+5)\} = -15; it was input into the virtual answer.

Kuri used social gestures in response to participants' answers in both conditions. For a correct answer, Kuri performed the positive physical gesture (``gotit''); for an incorrect answer, Kuri performed the negative physical gesture (``sad'').

In the Experiment condition, Kuri also used the positive and negative mixed reality arm gestures (Fig. \ref{fig:sixanims}) synchronized with the positive and negative physical gesture, respectively. We combined the physical and mixed reality gestures, as opposed to using mixed reality gestures only, based on feedback received from a pilot study. Participants in the pilot study indicated that gestures with both the body and mixed reality arms (as opposed to mixed reality arm gestures only) created a more integrated and natural robot appearance.

After the tutorial, participants attempted to solve a series of up to seven puzzles of increasing difficulty within a time limit of 10 minutes. When participants successfully input the correct answer to a puzzle, they advanced to the next puzzle. If the time limit was exceeded or all puzzles were completed, the system halted. Participants were then asked to do a survey and a semi-structured interview described in Section \ref{sec:measuresandanalysis}.

\subsection{Ensuring Gesture Presentation Consistency}

The puzzle task was designed to mitigate inconsistencies across participants. The first mitigation method addressed gesture randomness and diversity. The Experiment condition used gestures from the set of positive ($PG$ = \{cheer, clap, wave dance\}) and negative ($NG$ = \{facepalm, shoulder shrug, arm cross\}) gestures. To preserve balance, we first chose gestures randomly without replacement for each set, thereby guaranteeing that each gesture was shown, assuming at least 3 correct and 3 incorrect answers. Once all gestures from a group were shown, the gestures were chosen randomly, with replacement. The Control condition did not require methods for ensuring gesture diversity since Kuri used a single way of communicating correct answers and incorrect answers.

Steps were also taken to avoid only positive gestures being shown for users who had all correct answers. First, all participants were shown an incorrect answer and gesture during the tutorial. Second, some puzzles had a single physical block with two numbers on it (Fig. \ref{fig:twosideblocks}). In those cases, participants were told that the puzzle could have two answers. If their first answer was incorrect, they were told to turn the block over and use the number on the other side. Puzzles 3-7 all had this feature. Regardless of the participant's initial guess for these puzzles, they were told told they were incorrect and then shown a negative gesture. If the initial guess was one of the two possible answers, it was removed from the possible answers. After the initial guess, guesses were said to be correct if they were in the remaining set of correct answers. This consistency  method ensured that each participant saw all of the negative gestures.

\subsection{Part 2: Gesture Annotation} \label{sssec::task2}

All participants were shown a video of Kuri using the arm gestures, as seen in Fig. \ref{fig:sixanims} and can be found at \url{https://youtu.be/Ff08E9hvvYM}. The video was recorded through the Hololens camera, giving the same view as seen by participants in the Experiment condition of the math puzzles. After the participants watched all gestures once, they were given the ability to rewind and re-watch gestures as they responded to a survey. The gesture order of presentation was initially randomly generated and then presented in that same order to all participants. In total, the second part of the experiment took 5-10 minutes.

\subsection{Measures and Analysis} \label{sec:measuresandanalysis}

We used a combination of objective and subjective measures to characterize the difference between the conditions. 

\textbf{\textit{Task Efficiency}} was defined as the total time taken to complete each puzzle. We also noted users that did not complete all puzzles within the 10 minute time limit. The post-study 7-point Likert scale questions used 4 subjective measures, adapted from Heernik et al. \cite{heerink2010assessing} to evaluate the use of ARHMD with Kuri. The measures were: Total Social Presence, Attitude Towards Technology, Perceived Ease of Use, and Perceived Usefulness. \textbf{\textit{Total Social Presence}} measured the extent the robot was perceived as a social being (10 items, Cronbach's $\alpha=.89$). \textbf{\textit{Attitude Towards Technology}} measured how good or interesting the idea of using mixed reality with the robot was (3 items, Cronbach's $\alpha=.97$). \textbf{\textit{Perceived Ease of Use}} measured how easy the robot with mixed reality was to use (5 items, Cronbach's $\alpha=.73$). \textbf{\textit{Perceived Usefulness}} measured how useful or helpful the robot with mixed reality seemed (3 items, Cronbach's $\alpha=.81$).

Participants rated the robot's physical (0.00) to virtual (1.00) teammate presence to a granularity of two decimal points (e.g., 0.34) and were able to see and check the exact value they input. This measure was used to gauge where Kuri was perceived as a teammate on the Miligram virtuality continuum \cite{milgram1995telerobotic}. 

Qualitative coding was performed on the responses to the post-study semi-structured interviews, to assess how emotional and helpful Kuri seemed to the participants. Participants from the Experiment condition were also asked how ``attached'' the arms felt on Kuri; this question was coded for only those participants (Table \ref{table:interview}). To construct codes and counts, one research assistant coded for: ``How emotional was Kuri?'' and ``How helpful was Kuri?'' without looking at the data from the Experiment condition. Another assistant coded for: ``Do the arms seem to be a part of Kuri?'' for participants in the Experiment condition. Codes were constructed by reading through interview transcripts and finding ordinal themes. Example quotes for each code are shown in Table \ref{table:interview}.

For the gesture annotation, we used a similar approach to Marmpena et al. \cite{marmpena9does}: users annotated each robot gesture on a slider from very negative (-1.00) to very positive (+1.00), in order to measure valence. The slider granularity was to two decimal points (e.g. -0.73) and participants were able to see the precise decimal value they selected. 

To test annotator repeatability and ability to distinguish gestures, we conducted an inter-rater reliability test. We were interested in measuring the repeatability of choosing a single person from a generalized population to rate each gesture. To measure inter-rater reliability, we used intraclass correlation with a 2-way random effect model for a single participant against all participants (referred to as ``Raters'') among the six gestures (referred to as ``Subjects'') to find a measure for absolute agreement among participants. We used Eq. \ref{eq:ICC2} where $k$ denotes the number of repeated samples, $MS_R$ is the mean square for rows, $MS_E$ is the mean square error, $MS_C$ is the mean square for columns, and $n$ is the number of items tested \cite{koo2016guideline}. We used $k$ = 1 as we were interested in the reliability of agreement when choosing a single rater to rate the gestures against all other raters. We used the \textit{icc} function from the \textit{irr} package of R (v3.6.0, \url{https://cran.r-project.org/}) with parameters ``twoway'', ``agreement'', ``single''.  According to Koo et. al \cite{koo2016guideline}, poor values $< 0.5$, moderate values $<0.7$, good values $<0.9$, and excellent values $\geq 0.9$.  

\begin{equation} \label{eq:ICC2}
    ICC(2,k) = \frac{MS_R - MS_E}{MS_R+\frac{MS_C-MS_E}{n}}
\end{equation}

Each gesture also had an open-ended text box where users were asked: ``Please describe what you believe gesture X conveys to you'' where `X' referred to the gesture number. These textual data were later coded by a research assistant (Table \ref{table:qualgestuers}). Codes were constructed as the most common and relevant words for each gesture. Example quotes for each code are also included in Table \ref{table:qualgestuers}.

\section{Results}
\subsection{Participants}
A total of 34 participants were recruited and randomly assigned to one of two groups: Control (5F, 12M) and Experimental (8F, 9M).  Participants were University of Southern California students with an age range of 18-28 ($M=22.3, SD = 2.5$).  

\subsection{Arms Vs. No Arms Condition}
For the math puzzles, we analyzed our performance metric but saw no statistically significant effect between conditions. An independent-samples t-test was conducted to compare \textbf{\textit{Task Efficiency}} between the two experiment conditions. There was not significant difference in scores for arms ($M=83.0, SD=33.0$) and no arms ($M=77.8, SD=23.7$) conditions ($t(16), p=.54$). There were an equal number of participants (6) in each group who timed out at 10 minutes.

\begin{figure}[t]
  \centering
  \includegraphics[width=\columnwidth, height=7.0cm]{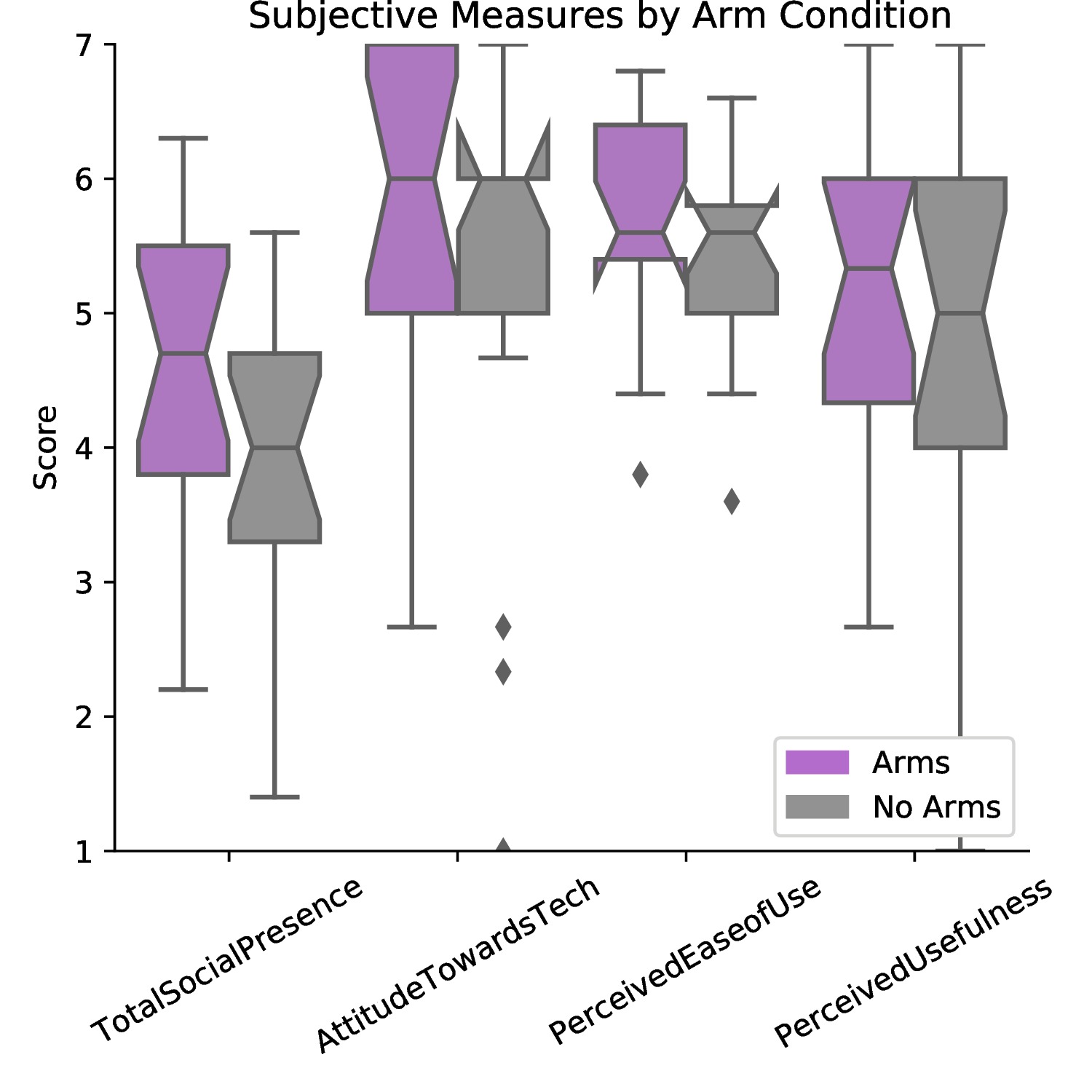}
  \caption{No statistical significance found for subjective measures. Boxes indicate 25\% (Bot), 50\% (Mid), and 75\% (Top) percentiles. Notches indicate the 95\% confidence interval about the median calculated with bootstrapping 1,000 particles \cite{efron1986bootstrap}. Thus notches can extend over the percentiles and give a ``flipped'' appearance (e.g., \{Attitude, NoArms\}).}
  \label{fig:conditionsubject}
\end{figure}

We saw no significant effect among each metric, as seen in Fig. \ref{fig:conditionsubject}. Mann-Whitney tests indicated no significant increases in \textbf{\textit{Total Social Presence}} between arms $(Mdn=4.7)$ and no arms $(Mdn=4.0)$ conditions ($U=115.5, p=.16$), \textbf{\textit{Attitude Towards Technology}} between arms $(Mdn=6.0)$ and no arms $(Mdn=6.0)$ conditions ($U=117.5, p=.18$), \textbf{\textit{Perceived Ease of Use}} between arms $(Mdn=5.6)$ and no arms $(Mdn=5.6)$ conditions ($U=117.0, p=.17$), and \textbf{\textit{Perceived Usefulness}} between arms $(Mdn=5.33)$ and no arms $(Mdn=5.0)$ conditions ($U=138.0, p=.41$). Qualitative coding for interviews can be found in Table \ref{table:interview}. An explanation of the qualitative coding used for the interviews is found in Section \ref{sec:measuresandanalysis}.

\begin{figure}[b]
  \centering
  \includegraphics[width=\columnwidth, height=5.5cm]{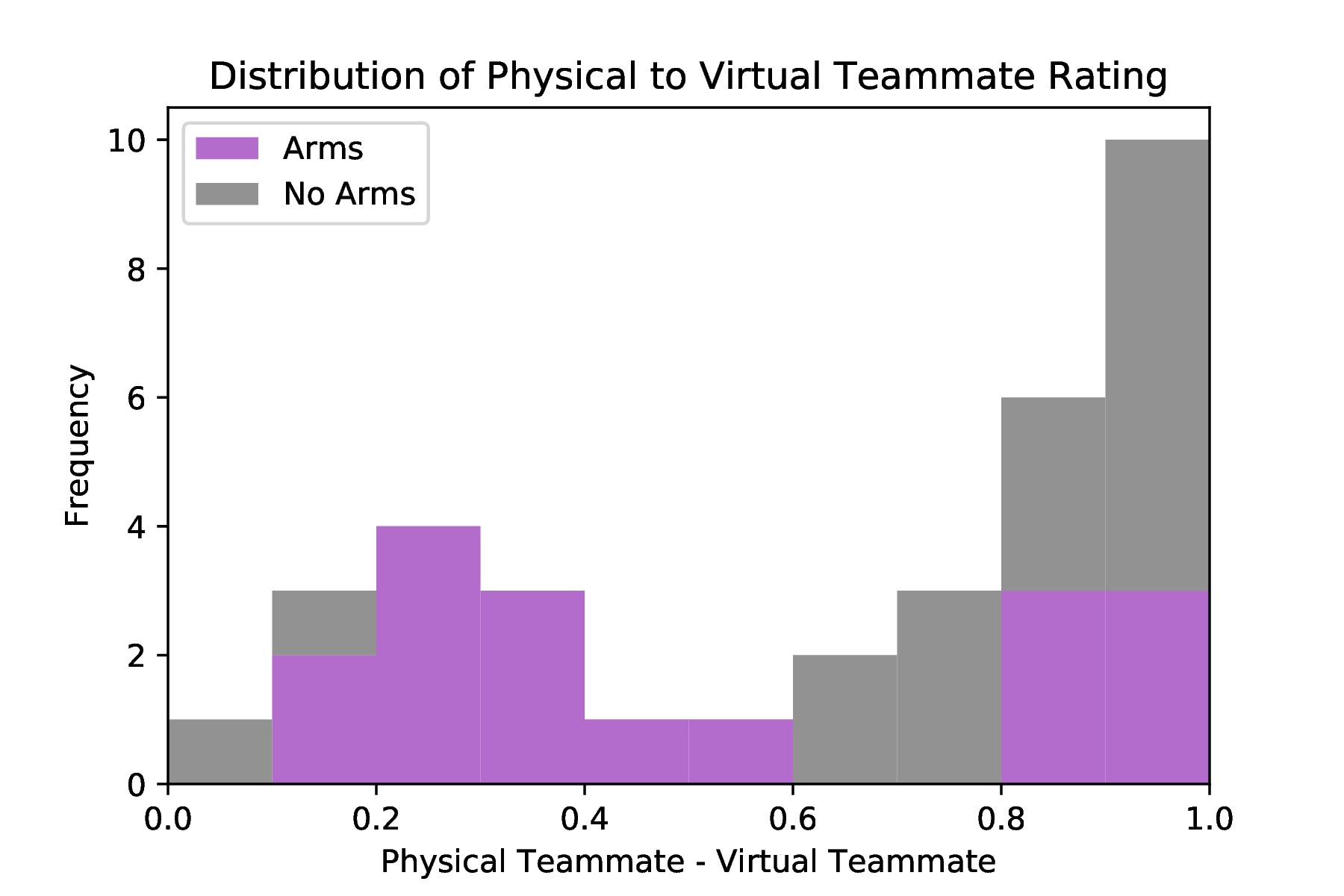}
  \caption{Stacked histogram with clustering to the left and right of 0.5 rating.}
  \label{fig:physvirthist}
\end{figure}

\begin{figure}[b]
  \centering
  \includegraphics[width=\columnwidth, height=5.5cm]{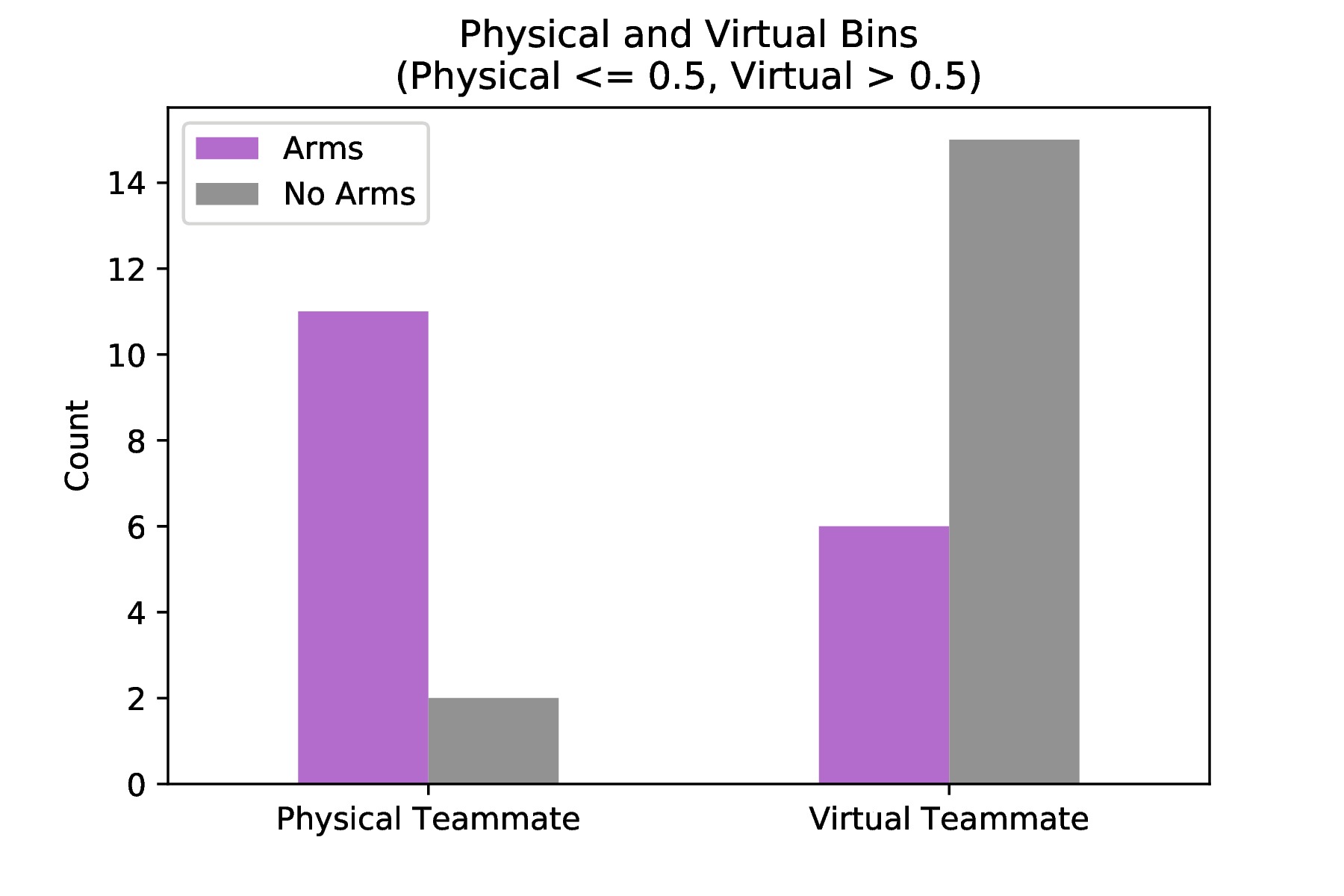}
  \caption{Participants in the Experiment condition were more likely to rate the mixed reality robot as physical.}
  \label{fig:physvirtbin}
\end{figure}

Most participants answered towards the ends of the physical-to-virtual teammate scale, with very few near the middle (Fig. \ref{fig:physvirthist}). Consequently, we divided the participants into two groups: ``Physical Teammate'' (ratings $\leq 0.5$, $n=13$)  and ``Virtual Teammate'' (ratings $>0.5$, $n=21$) (Fig. \ref{fig:physvirtbin}) and performed a Chi-Square Independence test. A significant interaction was found ($\chi^2(1),p=.002$). Participants in the Experiment condition, who experienced the arms, were more likely ($64.7\%$) to rate Kuri as a physical teammate than participants in the Control condition, who did not experience the arms ($11.8\%$). Next, we performed post-hoc analyses on subjective measures with the physical and virtual teammate binned groups.

\begin{table}[t]
\centering
\caption{Qualitative Interview Coding}
\label{table:interview}
\rowcolors{2}{white}{gray!25}
\begin{tabularx}{\columnwidth}{|>{\hsize=0.7\hsize}X|
                                >{\hsize=0.3\hsize}Y
                                >{\hsize=0.3\hsize}Y
                                >{\hsize=2.7\hsize}Y|}
\hline
Code & No Arms & Arms & Quote \\
\hline
Not Emotional      & 7   & 5      & ``I didn't feel any emotion from the robot''  \\
Close to Emotional & 9   & 7      & ``Like not so emotional because the task was not based on the emotion'' \\
Emotional          & 1   & 4      & ``It can talk and tell different emotions when I answer questions differently'' \\
Very Emotional     & 0   & 1      & ``When it went like *crosses arms* it was like `come on you're not helping me here.' And when her *acts out cheering*, yeah I would say very''\\
\hline
\hline
Not Helpful      & 6       & 2    & ``No'' \\
Somewhat Helpful & 2       & 3    & ``Sort of, yeah''\\
Helpful          & 9       & 12   & ``I like the way it had the visual feedback when I get right or wrong, and I just feel like it could reinforce it."\\
\hline
\hline
Are Arms a Part of Kuri? & - & Arms Count & Quote  \\
\hline
No                   & - & 2     & ``They seemed pretty detached''  \\
Somewhat             & - & 4     & ``When it was pointing things it did seem like it a little bit'' \\
Mostly               & - & 3     & ``I would say 60 percent'', ``8/10'' \\
Yes                  & - & 8     & ``What gave me the most information was her arms'' \\ 
\hline
\end{tabularx}
\end{table}

\begin{figure}[t]
  \centering
  \includegraphics[width=\columnwidth, height=7cm]{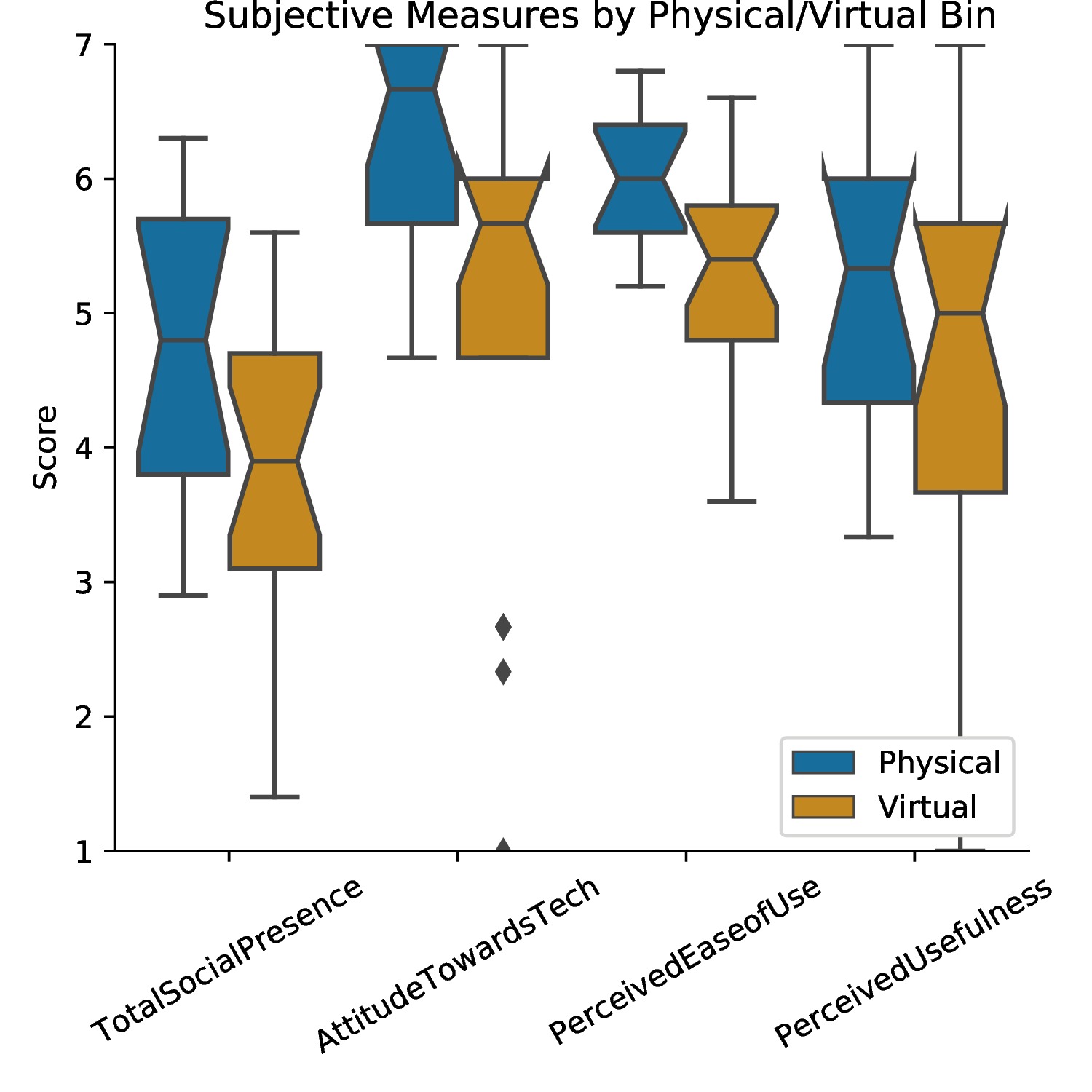}
  \caption{Significant increases for the first 3 measures with a marginally significant increase for measure 4. See Fig. \ref{fig:conditionsubject} for notch box-plot explanation.}
  \label{fig:physsubject}
\end{figure}

\subsection{Physical Vs. Virtual Teammate Bins}
We analyzed our survey data with regard to the two bins and saw significant effects among metrics (Fig. \ref{fig:physsubject}). Mann-Whitney tests indicated a significant increase in \textbf{\textit{Total Social Presence}} between physical $(Mdn=4.8)$ and virtual $(Mdn=3.9)$ groups ($U=86.5, p=.04, \eta^2=.092$), \textbf{\textit{Attitude Towards Technology}} between physical $(Mdn=6.7)$ and virtual $(Mdn=5.7)$ groups ($U=73.0, p=.01, \eta^2=.149$), and \textbf{\textit{Perceived Ease of Use}} between physical $(Mdn=6.0)$ and virtual $(Mdn=5.4)$ groups ($U=79.0, p=.02, \eta^2=.122$). We found only a marginal significant increase for \textbf{\textit{Perceived Usefulness}} between physical $(Mdn=5.3)$ and virtual $(Mdn=5.0)$ groups ($U=93.5, p=.07, \eta^2=.068$).

\begin{table*}[t]
\centering
\caption{Gesture Description Qualitative Code Counts}
\label{table:qualgestuers}
\rowcolors{2}{white}{gray!25}

\begin{tabularx}{\textwidth}{|>{\hsize=0.4\hsize}X|>{\hsize=1.0\hsize}Y>{\hsize=1.6\hsize}Y|}
\hline
\textbf{Gesture}    & \textbf{Code : Count}                                  & \textbf{Example Quote}                                                                                       \\
\hline
G1: Facepalm   & Disappointment :10, Frustration: 4, Facepalm: 3 & ``Facepalm, the robot is frustrated/disappointed''                                                  \\
G2: Cheer     & Happy: 8, Celebration: 7, Cheer: 6                 & ``That you got the answer correct and the robot is cheering you on''                                  \\
G3: Shrug      & Don't Know Answer: 11, Shrug: 5, Confuse: 4       & ``Shrugging, he doesn't know what the person is doing or is disappointed in the false guess''       \\
G4: Arm Cross  & Angry: 8, Disappointment: 7, Arm Cross: 4          & ``Crossing arms. \emph{`Really??'} mild exasperation or judgment.''                                          \\
G5: Clap      & Happy: 13, Clapping: 7, Excited: 4                 & ``It's a very happy, innocent clap. I like the way its eyes squint, gives it a real feeling of joy.'' \\
G6: Wave Dance & Happy: 11, Celebrate: 4, Good Job: 4               & ``Celebration dance, good job!''                     \\
\hline
\end{tabularx}
\end{table*}

\begin{table}[b]  
\centering
\caption{Valence Rating Percentiles by Gesture.}
\label{table:gesturemeds}
\rowcolors{2}{white}{gray!25}
\begin{tabularx}{\columnwidth}{|X|YYYYYY|}
\hline
 \textbf{$P_\%$} & \textbf{G1}  & \textbf{G2}  & \textbf{G3} & \textbf{G4}  & \textbf{G5} & \textbf{G6} \\
\hline
$P_{25}$     & -0.85 & 0.72 & -0.59 & -0.87 & 0.64 & 0.55 \\
$P_{50}$     & -0.69 & 0.88 & -0.30 & -0.58 & 0.84 & 0.76 \\
$P_{75}$     & -0.41 & 1.00 & -0.09 & -0.36 & 1.00 & 1.00 \\
\hline
\end{tabularx}
\end{table}

\begin{figure}[t]
  \centering
  \includegraphics[width=\columnwidth]{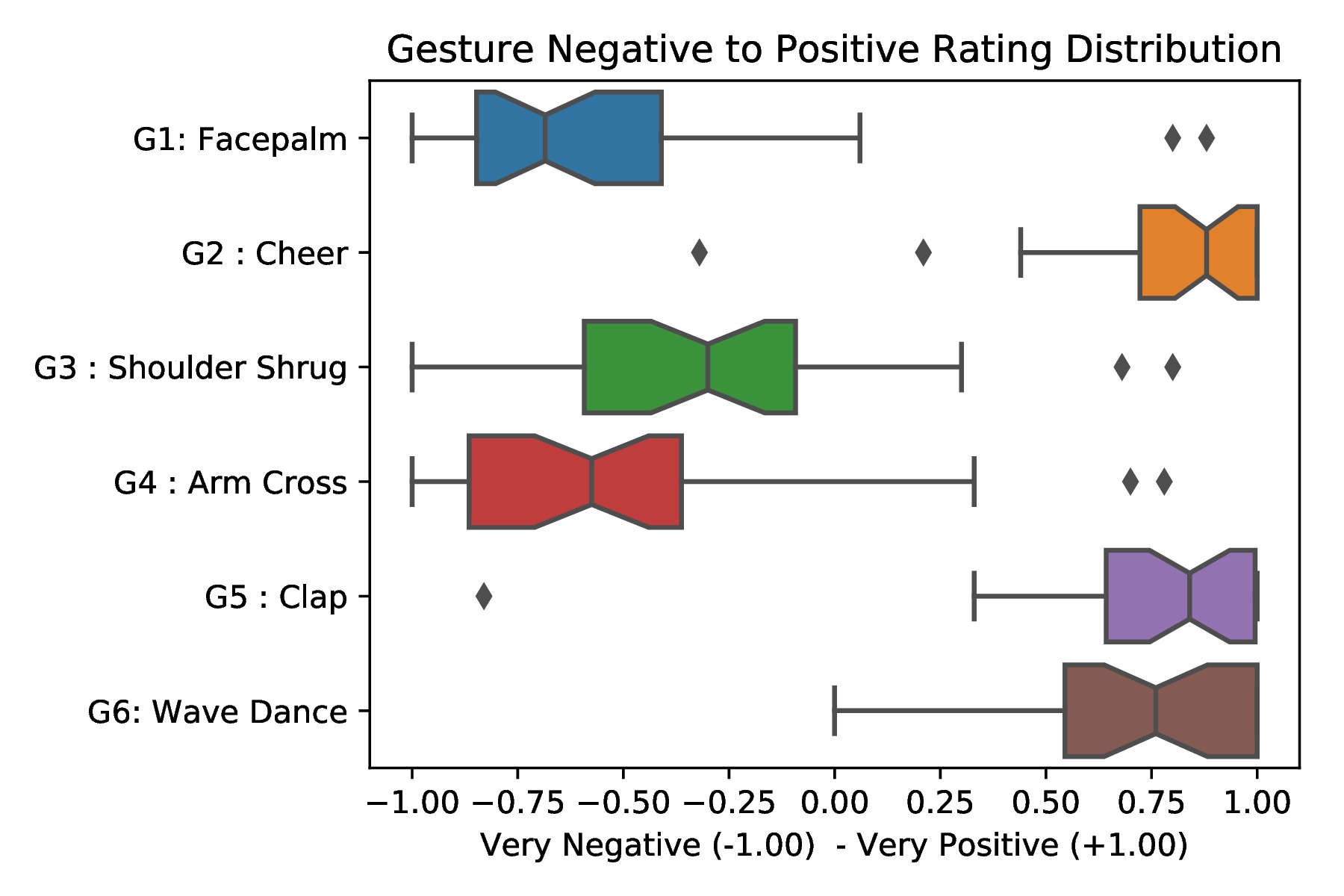}
  \caption{Distribution on ability to differentiate gesture valence.}
  \label{fig:gesturebox}
\end{figure}

\subsection{Gesture Validation}
We analyzed the data from gesture annotation in order to validate participants' ability to distinguish the valence of gestures and consistency in interpreting gestures. As seen in Fig. \ref{fig:gesturebox} and Table \ref{table:gesturemeds}, participants could distinguish the valence (negativity to positivity) of the gestures. The two-way, agreement intraclass correlation for a single rater, described in Section \ref{sec:measuresandanalysis}, resulted in a score $ICC(A,1)$ = 0.77  with 95$\%$ confidence interval 0.55-0.95, and $F(5,190)$ = 125, $p < 0.001$, which constitutes moderate to good reliability. Qualitative data are summarized in Table \ref{table:qualgestuers}. Explanation for coding these data can be found in Section \ref{sec:measuresandanalysis}.

\section{Discussion}

The arms vs. no arms conditions did not show statistical significance for \textbf{\textit{Task Efficiency}} nor for subjective measures. However, the two conditions were highly correlated with user perception of either a physical or virtual teammate as binned categories. We postulate that participants may have associated arms in general with more physical tasks, such as picking up objects or pointing. The Experiment condition also involved more overall movement, which may have conveyed Kuri as more of a physical teammate to participants who may have associated movement with physicality. 

The binned subjective results suggest a "better" teammate for a physically associated mixed reality robot. This is consistent with the evidence for the importance of embodiment for social presence \cite{deng2019embodiment}. Although having arms strongly correlated with the perception of a physical teammate, there could be other factors that influenced physical presence. Future mixed reality robot research may explore factors that increase physical presence of the overall agent to ground mixed robot abilities such as the increased social expressive range of gestures discussed in this work. Given the flexibility and the lack of physical constraints of the MR arm interface, new gestures and actions could be added and adapted to other scenarios, as explored in some previous work in AR-HRI \cite{walker2018communicating, williams2019virtual}.

Gestures were distinguishable on a valence scale with a very high agreement reliability for the scoring among participants, as seen in the intervals reported in Table \ref{table:gesturemeds}. This suggests gesture annotation is highly repeatable. Users had slightly more difficulty in rating the negative gestures than positive ones. Two participants also rated all gestures as positive, as indicated by the 6 erroneous marks for the 3 negative gestures in Fig. \ref{fig:gesturebox}. These data were included in all reported statistics and may indicate a confusion in the rating scale. The qualitative data (Table \ref{table:qualgestuers}) also support the distinguishability of the gestures as intended (i.e., indicating it was a ``clapping'' gesture).

\section{Limitations and Future Work}

The first generation Hololens posed many issues. As reported by participants, the virtual arms were difficult to see due to the limited field of view of the AR/VR headset. Hand tracking could be used to supplement direct, intuitive, and safe interaction with mixed reality extensions to robots. For example, the mixed reality arms in this study could be given the ability to share a high-five gesture with a study participant. Eye tracking, which has been used for modeling engagement \cite{rich2010recognizing} and joint attention \cite{warren2015can}, can also provide real time input towards autonomous control for prompting users or to gesture to points of interest.

The reported study consisted of a limited number of gestures and limited measures of those gestures. Although results showed that the gestures were distinguishable on a scale for valence and their intended meaning, we did not have subjects annotate for arousal. The two-factor scale of valence and arousal is commonly used as a metric of measuring affect and has been used previously to rate robot gestures \cite{marmpena9does}. Measuring arousal or a higher feature scale for gestures could provide further insight into their perceived meaning and into how they differ from physical gestures.

Many study participants reported in their post-study interviews that Kuri's interactions were limited.  Reports of Kuri being seen as a ``referee'' suggest that the robot is seen as closer to a judge than a peer or teammate. Adaptive robot characters could potentially leverage the unique modalities, constraints, and data ARHMD provide in order to more adaptively meet user preferences, leveraging the socially expressive capabilities of the mixed reality arms.

\addtolength{\textheight}{-0cm}   




\section{CONCLUSION}
This work explored the use of mixed reality arms for increasing the range of social expressivity for low-expressivity robots. Integration of the expressive modality of mixed reality has the potential to increase a robot's expressive range as well as increase its physical presence. Future work may explore increasing the expressive range while also leveraging real-time data (e.g., eye gaze) to estimate user engagement and express appropriate mixed reality robot responses. This could lead toward more fluid, expressive, and effective human-robot interaction.

\section{ACKNOWLEDGMENT}
We would like to thank Stefanos Nikolaidis, Matthew Reuben, and Jessica Lupanow for all of their assistance.

\bibliographystyle{ieeetr}
\bibliography{bibliograph}

\end{document}